%% file: main.tex
\documentclass{article}

\usepackage{microtype}
\usepackage{graphicx}
\usepackage{subfigure}
\usepackage{booktabs} 

\usepackage{hyperref}

\usepackage[accepted]{icml2019_phys4dl}

\usepackage{color}
\definecolor{royalblue}{rgb}{0.0, 0.14, 0.4}
\hypersetup{citecolor=royalblue}

\usepackage{amsmath}
\usepackage{amsfonts}
\usepackage{tabularx}
\usepackage[algoruled,algo2e]{algorithm2e}
\usepackage[percent]{overpic}
\newcommand\Ell{\mathcal L}

\newcommand\bX{\mathbf X}
\newcommand\bZ{\mathbf Z}

\newcommand\bg{\boldsymbol\Gamma}

\newcommand\IR{\mathbb R}

\begin{document}

\twocolumn[
\icmltitle{A Quantum Field Theory of Representation Learning}



\icmlsetsymbol{equal}{*}

\begin{icmlauthorlist}
\icmlauthor{Robert Bamler}{uci}
\icmlauthor{Stephan Mandt}{uci}
\end{icmlauthorlist}

\icmlaffiliation{uci}{Donald Bren School, Department of Computer Science, University of California at Irvine, USA}

\icmlcorrespondingauthor{Robert Bamler}{rbamler@uci.edu}
\icmlcorrespondingauthor{Stephan Mandt}{mandt@uci.edu}

\icmlkeywords{Representation Learning, Time Series, Physics}

\vskip 0.3in
]



\printAffiliationsAndNotice{} 

\input{intro.tex}
\input{symmetry-breaking.tex}
\input{method.tex}
\input{experiments.tex}

\bibliography{references}
\bibliographystyle{icml2019}

\end{document}

%% file: intro.tex

\begin{abstract}
Continuous symmetries and their breaking play a prominent role in contemporary physics.
Effective low-energy field theories around symmetry breaking states explain diverse phenomena such as superconductivity, magnetism, and the mass of nucleons.
We show that such field theories can also be a useful tool in machine learning, in particular for loss functions with continuous symmetries that are spontaneously broken by random initializations.
In this paper, we illuminate our earlier published work~\citep{bamler2018improving} on this topic more from the perspective of theoretical physics.
We show that the analogies between superconductivity and symmetry breaking in temporal representation learning are rather deep, allowing us to formulate a gauge theory of `charged' embedding vectors in time series models.
We show that making the loss function gauge invariant speeds up convergence in such models.
\end{abstract}

\section{Introduction}
\label{sec:intro}

Continuous symmetries are of central interest in field theory.
Gauge fields mediate interactions by propagating continuous symmetries transformations across space.
Continuous symmetries are also common in representation learning, such as in word embeddings.
In fact, the phrase `dense embeddings' typically refers to models that do not distinguish any special directions in the representation space.
Since no direction is special and only relative orientations matter, dense embedding models are often invariant under an arbitrary simultaneous rotations of all embedding vectors.

For example, factorizing a large matrix $X\approx U^\top V$ into two smaller matrices $U$ and $V$ is invariant under the transformation $U\mapsto RU, V\mapsto RV$ with any orthonormal matrix~$R$ since $(RU)^{\!\top}\! (RV)=U^\top\! R^\top\! R \,V = U^\top V$. In this paper, we consider models with continuous symmetries of this kind.

In~\citep{bamler2018improving}, we recently showed that certain time series models with continuous symmetries suffer from slow convergence of Gradient Descent (GD). We proposed the Goldstone-GD algorithm, which speeds up convergence by using artificial gauge fields.
While our motivation for Goldstone-GD came from the theory of superconductivity, the paper glossed over this connection as it addressed a general machine learning audience with no physics background.

Given the topic of this workshop, this paper primarily addresses readers that have some familiarity with concepts of theoretical physics, allowing them to understand these analogies on a deeper level.
We expose a profound relation between the Goldstone-GD algorithm and gauge theory in general, and the theory of superconductivity in particular.

In Section~\ref{sec:goldstone}, we discuss the concept of symmetry breaking in physics and in representation learning models for time series.
The analogy shows that the considered models suffer from slow convergence due to a so-called gapless excitation spectrum, i.e., their loss function is ill-conditioned.
In Section~\ref{sec:method}, we show that the analogy goes even further in that the considered models are similar to superfluids like helium-4 at low temperatures.
This observation allows us to use a key insight:
as \emph{charged} superfluids (i.e., superconductors) have a gapped energy spectrum, we can solve the slow convergence problem by assigning a charge to embedding vectors, i.e., by coupling them to a gauge field.
Section~\ref{sec:experiments} repeats an experiment from~\citep{bamler2018improving}.

%% file: symmetry-breaking.tex

\section{Symmetry Breaking and Goldstone Modes}
\label{sec:goldstone}

In this section, we introduce core concepts of the theory of spontaneous symmetry breaking in physics (Subsection~\ref{sec:goldstone-physics}) and apply them to machine learning (Subsection~\ref{sec:goldstone-ml}).

\subsection{Goldstone Modes in Physics}

\label{sec:goldstone-physics}

We discuss the properties of Goldstone modes in physics.
In Section~\ref{sec:goldstone-ml}, we show that the same theory also applies to Markovian time series models in machine learning.

\paragraph{Example: Phonons.}
Atoms in a solid arrange in a periodic lattice.
This admits for a very compact description of the microscopic configuration:
to provide the positions of all atoms, we only need to specify a global offset position with respect to some reference lattice with the right periodicity.
This phenomenon is called spontaneous symmetry breaking, and the global offset is an example of an order parameter.
While the model (Hamiltonian) is invariant under arbitrary translations, the state breaks this continuous symmetry on a global level by picking a value for the order parmeter.

Real crystals are not perfectly periodic, and more realistic description therefore generalizes the order parameter to a smooth function of position and time.
Waves in the order parameter field are called Phonons in crystals, and Goldstone modes in general.
It turns out that Goldstone modes cost arbitrarily little energy in the long wavelength limit~\citep{altland2010condensed}.
For example, Figure~\ref{fig:phonon-dispersion}a shows a measured Phonon dispersion relation, i.e., the phonon energy~$E$ as a function of the wave vector $q=\frac{2\pi}{\text{wave length}}$.
We observe that~$E\to0$ for $q\to0$, i.e., for smooth waves with long wavelength.
One says that the phonon spectrum is `gapless', i.e., the energy gap between the ground state and the lowest excitation goes to zero as the size of the system grows.

\paragraph{Other Examples of Goldstone Modes.}
Goldstone modes are as ubiquitous as spontaneous continuous symmetry breaking.
For example, in a ferromagnet, the magnetization breaks rotational symmetry, and its Goldstone modes are called `magnons'.
In quantum chromo dynamics, `pions' result from the spontaneous breaking of an approximate chiral symmetry.
In superfluids, Goldstone modes arise due to spontaneous breaking of the $U(1)$ phase symmetry of quantum mechanics.
As we discuss in Section~\ref{sec:higgs}, \emph{charged} superfluids nevertheless have a gap in the energy spectrum due to the so-called Higgs mechanism.
It is this mechanism that motivated the algorithm presented in this work.

\subsection{Goldstone Modes in Time Series Models}
\label{sec:goldstone-ml}

We show that spontaneous continuous symmetry breaking arises in a general class of representation learning models for time series.
We proof that the loss function of these models is ill-conditioned due to the existence of Goldstone modes.
This subsection follows~\citep{bamler2018improving}.

\paragraph{Representation Learning for Time Series.}
We introduce a general class of representation learning models for time series.
Consider sequential data $\bX\equiv\{X_t\}_{t=1:T}$ over~$T$ time steps $t\in\{1,\ldots,T\}$.
For each time step~$t$, we fit a dense embedding model with parameters~$Z_t$ to the data~$X_t$ by minimizing some local loss function~$\ell(X_t; Z_t)$ over~$Z_t$.
In our notation, the model parameters $Z_t$ form a matrix whose columns are $d$-dimensional embedding vectors.
As discussed in the introduction, we assume that the local loss function is invariant under arbitrary rotations~$R$,
\begin{align} \label{eq:local-rotsym}
    \ell(X_t; R Z_t) &= \ell(X_t; Z_t) \qquad \forall R\in SO(d).
\end{align}

To share statistical strength across the time dimension, we add a quadratic Markovian (i.e., nearest neighbor) coupling with a strength~$\lambda$.
Denoting the concatenation of all model parameters as $\bZ\equiv \{Z_t\}_{t=1:T}$, the total loss function is thus
\begin{align} \label{eq:total-loss}
    \Ell(\bZ) &= \sum_{t=1}^T \ell(X_t; Z_t)
        + \frac{\lambda}{2} \sum_{t=2}^T  ||Z_t - Z_{t-1}||_2^2.
\end{align}

Eqs.~\ref{eq:local-rotsym}-\ref{eq:total-loss} define a general class of representation learning models for time series.
The model class includes, e.g., dynamic matrix factorizations \citep{lu2009aspatiot,koren2010collaborative,sun2012dynamic,charlin2015dynamic} and dynamic word embeddings \citep{bamler_dynamic_2017,rudolph2017dynamic}.

\begin{figure}[t]
    \vskip 0.04in
    \begin{center}
    \centerline{\includegraphics[width=\columnwidth]{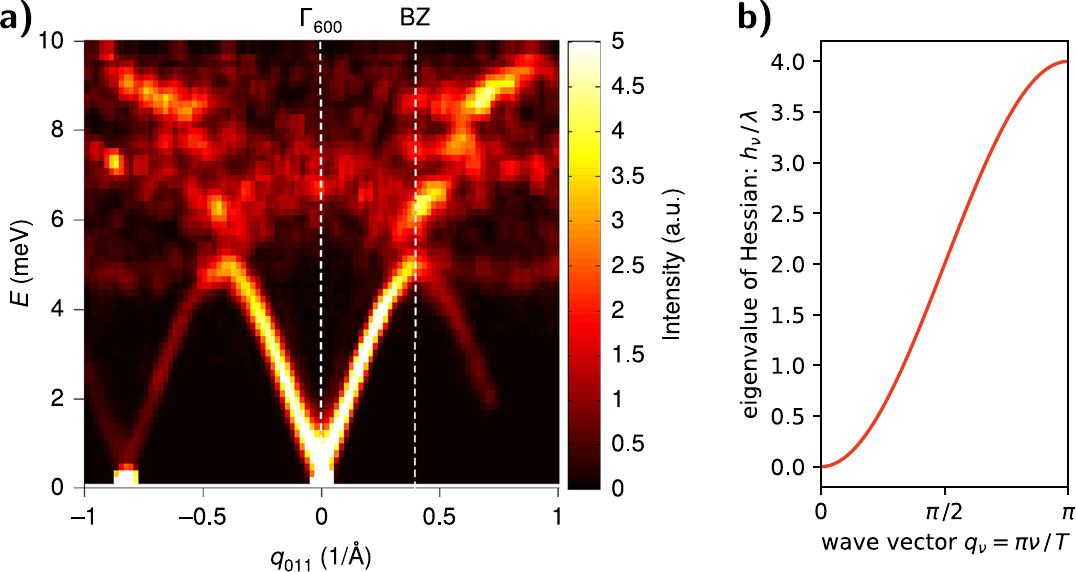}}
    \caption{Dispersion relations of Goldstone modes;
    a)~phonons in $\text{Ba}_{7.81}\text{Ge}_{40.67}\text{Au}_{5.33}$, measured with neutron scattering;
    b)~eigenvalues~$h_\nu$ of the Hessian of the regularizer in Eq.~\ref{eq:total-loss}, see derivation in Section~\ref{sec:goldstone-ml};
    both spectra are gapless, i.e., $E,h_\nu\to0$ as $q\to0$.
    This is a generic property of Goldstone modes.
    Figure a) taken from~\citep{lory2017direct} with the authors' permission; CC-BY.}
    \label{fig:phonon-dispersion}
    \end{center}
    \vskip -0.2in
\end{figure}

\begin{figure}[t]
    \vskip 0.04in
    \begin{center}
    \centerline{\includegraphics[width=\columnwidth]{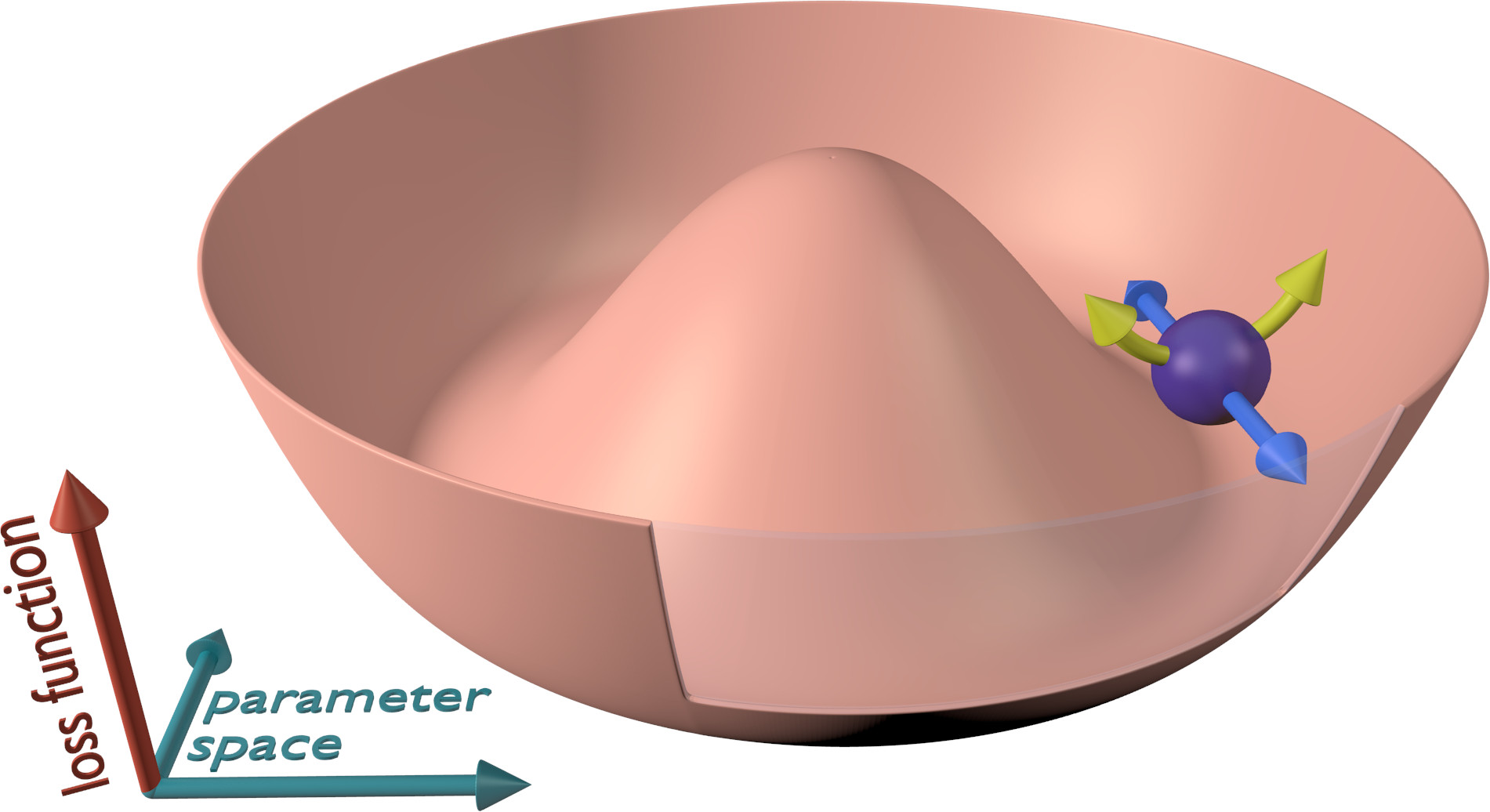}}
    \caption{A rotationally symmetric potential~$\ell$ has zero curvature within the symmetry subspace (blue arrows).}
    \label{fig:champagne_bottle}
    \end{center}
    \vskip -0.2in
\end{figure}

\paragraph{Goldstone Modes.}
We show that minimizing the loss function~$\Ell$ defined in Eqs.~\ref{eq:local-rotsym}-\ref{eq:total-loss} is an ill-conditioned optimization problem due to the existence of Goldstone modes.
The argument goes in two steps.
First, for each time step~$t$, the local loss function~$\ell$ has a manifold of degenerate minima:
if~$Z_t^*$ minimizes~$\ell(X_t;\,\cdot\,)$, then, according to Eq.~\ref{eq:local-rotsym}, so does $R Z_t^*$ for any $R\in SO(d)$.
The gradient $\nabla_{\!Z_t}\ell(X_t;Z_t)$ is zero and therefore constant over the entire manifold of degenerate minima, implying that the Hessian (i.e., the second derivative) has zero eigenvalues within the subspace spanned by infinitesimal local rotations~(blue arrows in Figure~\ref{fig:champagne_bottle}).
Thus, within this subspace, only the Hessian~$\mathbf H_\text{reg}$ of the regularizer term in Eq.~\ref{eq:total-loss} contributes to the Hessian of~$\Ell$.

Second, $\mathbf H_\text{reg}$ is ill-conditioned.
The Markovian regularizer term in Eq.~\ref{eq:total-loss} has the form of the potential energy of a chain of~$T$ springs with spring constant~$\lambda$.
The eigenmodes of such a chain are harmonic waves $\propto\cos(q_\nu (t-\frac12))$ with $q_\nu =\pi\nu /T$ for $\nu\in\{0,\ldots,T-1\}$, and with eigenvalues~$h_\nu=(2-2\cos(q_\nu))\lambda$, see Figure~\ref{fig:phonon-dispersion}b.
This is a \emph{gapless} spectrum, i.e., in long time series (large~$T$), the lowest eigenvalues become arbitrarily small:
$h_\nu = O(\lambda/T^2)$ for small~$\nu$.
This leads to a large condition number $h_{T-1}/h_1 = O(T^2)$ and therefore to slow convergence of gradient descent (GD).

%% file: method.tex

\section{Goldstone Gradient Descent}
\label{sec:method}

We propose a solution to the slow convergence problem from Section~\ref{sec:goldstone-ml}, motivated by an analogy to the quantum field theory of superconductivity.
We discuss the physical intuition in Subsection~\ref{sec:higgs} and apply it to time series models in Subsection~\ref{sec:algorithm}.
This work focuses on the fundamental physical principles invoked by the algorithm.
A more pragmatic exposition with pseudocode instead of physical interpretations was given in~\citep{bamler2018improving}.

\subsection{Higgs Mechanism in Superconductivity}
\label{sec:higgs}

Our intuition for the algorithm described in Section~\ref{sec:algorithm} comes from comparing uncharged and charged superfluids.
This section provides a very compact summary of these theories;
for more details see, e.g.,~\citep{altland2010condensed}.

In quantum mechanics, the global phase of a wave function cannot be observed, implying a $U(1)$ symmetry.
The ground state of an electrically neutral superfluid, e.g., helium-4 at low temperatures, breaks this symmetry spontaneously, leading to a Goldstone mode and a gapless excitation spectrum.

The situation is different in electrically charged superfluids, i.e., superconductors.
For charged particles, the global $U(1)$ symmetry is promoted to a local gauge symmetry
as the gradient of the wave function couples to the electromagnetic gauge fields
(details see `minimal coupling' in Section~\ref{sec:algorithm} below).
The Goldstone mode inherits this coupling, leading to the Anderson-Higgs mechanism:
the gauge field acquires a mass by `consuming' the Goldstone mode.
The photon mass makes magnetic fields energetically expensive so that the superconductor expels magnetic fields from its interior (`Meissner effect').
A quantitative derivation of the decay of magnetic fields at the surface of a superconductor requires considering the \emph{dynamics} of the gauge fields.

\subsection{Fast Optimization with Charged Embeddings}
\label{sec:algorithm}

We propose an optimization method that quickly eliminates Goldstone modes, thereby solving the slow convergence problem discussed in Section~\ref{sec:goldstone-ml}.
We follow three steps.
First, using the physical intuition from Section~\ref{sec:higgs}, we give the embeddings~$\bZ$ a \emph{charge}, i.e., we couple them to an auxiliary gauge field.
Second, in analogy to our comments on the Meissner effect, we introduce a dynamics for the auxiliary gauge fields.
Third, we apply a gauge transformation to turn the decay of gauge fields into a decay of Goldstone modes.

\paragraph{Minimal Coupling.}
Quantum mechanics encodes a state in a complex valued wave function $\psi(\mathbf r,t)$, where~$\mathbf r$ and~$t$ are space and time coordinates, respectively.
The theory is invariant under a global~$U(1)$ symmetry that rotates the wave function by an arbitrary phase.
However, it is not invariant under \emph{local} phase rotations that map $\psi(\mathbf r, t)$ to $\tilde\psi(\mathbf r, t) := e^{i\varphi(\mathbf r, t)}\psi(\mathbf r,t)$ with a local phase $\varphi(\mathbf r,t)\in\IR$.

For electrically charged particles, the global~$U(1)$ symmetry is promoted to a local gauge symmetry by coupling~$\psi(\mathbf r,t)$ to so-called gauge fields $V(\mathbf r,t)$ and $\mathbf A(\mathbf r,t)$.
The couplings are such that all observable effects of the above local phase rotation are compensated if we also change the gauge fields to $\tilde V := V - \nabla_{\!t}\varphi$ and $\mathbf{\tilde A} := \mathbf A + \nabla_{\!\mathbf r}\varphi$, respectively.
This is called a gauge transformation.

We adapt the concept of gauge invariance to the time series models from Section~\ref{sec:goldstone-ml}.
Due to Eq.~\ref{eq:local-rotsym} and the isotropic regularizer, the loss~$\Ell$ in Eq.~\ref{eq:total-loss} is globally $SO(d)$ symmetric,
\begin{align} \label{eq:global-sym}
  \Ell(RZ_1,\ldots,RZ_T) &= \Ell(Z_1,\ldots,Z_T) \;\; \forall R\in SO(d).
\end{align}
We elevate this global symmetry to a local gauge symmetry by introducing $t$-dependent rotation matrices $R_t\in SO(d)\;\forall t\in\{1,\ldots,T\}$.
Similar to how we expressed local phase rotations above as $e^{i\varphi(\mathbf r,t)}$, we can enforce the constraint $R_t\in SO(d)$ by parameterizing~$R_t$ as the matrix exponential of a skew symmetric matrix $\Gamma_{\!t}=-\Gamma_{\!t}^\top$,
\begin{align} \label{eq:matrix-exp}
  R_t := e^{\Gamma_{\!t}} = I + \Gamma_{\!t} + \frac12 \Gamma_{\!t}^2 + \frac{1}{3!} \Gamma_{\!t}^3 + \ldots \in SO(d)
\end{align}
Using the shorthand $\bg \equiv \{\Gamma_{\!t}\}_{t=1:T}$, we define the gauge invariant loss function,
\begin{align} \label{eq:gaugesym-loss}
  \Ell'(\bZ; \bg) := \Ell(R_1 Z_1, \ldots, R_T Z_T).
\end{align}
By construction, $\Ell'$ is invariant under gauge transformations.
Such a gauge transformation changes~$\Gamma_{\!t}$ to any other skew symmetric matrix $\tilde\Gamma_{\!t}$, and $Z_t$ to $\tilde Z_t := e^{-\tilde\Gamma_{\!t}} e^{\Gamma_{\!t}} Z_t$.
We thus say that the embeddings~$\bZ$ acquire a charge, and we call the components of~$\bg$ `gauge fields'.%
\footnote{Strictly speaking, gauge fields are finite differences of~$\Gamma_{\!t}$'s.}
Note that $e^{-\tilde\Gamma_{\!t}} e^{\Gamma_{\!t}}$ can not be simplified to $e^{\Gamma_{\!t}-\tilde\Gamma_{\!t}}$ since $SO(d)$ is a non-abelian group.
In this regard, the model is more similar to quantum chromodynamics than it is to quantum electrodynamics.

\paragraph{Minimization over Gauge Fields.}
The next step is to introduce a dynamics for the gauge fields~$\bg$.
To find optimal rotations~$R_t$, we minimize~$\Ell'$ over~$\bg$ for fixed $\bZ$.
We thus insert Eqs.~\ref{eq:matrix-exp}-\ref{eq:gaugesym-loss} into the loss function, Eq.~\ref{eq:total-loss}.
Because the local loss functions~$\ell$ are rotationally symmetric (Eq.~\ref{eq:local-rotsym}), they are independent of~$\bg$ and we only have to minimize the regularizer term.
We truncate~$\Ell'$ after the quadratic term in~$\bg$, resulting in the objective function
\begin{align} \label{eq:def-ell2prime}
  \sum_{t=2}^T \,{\rm Tr}\!\left[
      \left(\Gamma_{\!t} - \Gamma_{\!t-1} - \frac12 (\Gamma_{\!t} - \Gamma_{\!t-1})^2  \right) Z_{t-1} Z_t^\top \right]
\end{align}
where the $d\times d$ matrices $Z_{t-1} Z_t^\top$ can be precalculated at the beginning of the optimization.
The truncation after the quadratic term in~$\bg$ is asymptotically exact as, due to the gauge transformation described in the next paragraph, Goldstone modes decay over the course of the minimization and the minimum~$\bg^*$ of~$\Ell'$ therefore approaches $\bg^* \to \mathbf 0$.

The minimization over~$\bg$ is in principle equally ill-conditioned as the original minimization problem.
However, this is not an issue in practice.
First, Eq.~\ref{eq:def-ell2prime} is an explicit quadratic form in~$\bg$.
We can therefore minimize it efficiently either by solving a system of linear equations, or, if this is too expensive, by using GD with full-rank preconditioning (`natural gradients').
Second, such specialized optimization methods are computationally possible because the optimization over~$\bg$ is over a much lower dimensional space than that over~$\bZ$:
$\bg$ lives entirely in the embedding space, i.e., unlike~$\bZ$, its matrix dimensions are independent of the number of embedding vectors.

\paragraph{Gauging Away Goldstone Modes.}
Minimizing over~$\bg$ reduces the gauge invariant loss function~$\Ell'(\bZ;\bg)$ but it leaves the embeddings~$\bZ$ and therefore the original loss~$\Ell(\bZ)$ invariant.
The final step of the algorithm is to apply a gauge transformation that reduces the value of~$\Ell(\bZ)$ by eliminating Goldstone modes.
We update the embedding vectors as $Z_t \gets R_t Z_t = e^{\Gamma_{\!t}}Z_t$.
As in the optimization phase, truncating the matrix exponential function to a finite order leads to an asymptotically correct result as~$\bg\to\mathbf0$.

\paragraph{Overall Training Loop.}
The gauge transformation minimizes~$\Ell$ only along the directions of local rotations.
Since the local loss functions~$\ell$ are invariant under local rotations, they remain unchanged under gauge transformations.
To optimize over the entire embedding space, we alternate between a phase of the above gauge field optimization, and a phase of standard gradient descent on~$\Ell$ over~$\bZ$.
This concludes the Goldstone-GD optimization algorithm.

%% file: experiments.tex

\section{Experiment}
\label{sec:experiments}

\begin{figure}[t]
    \vskip 0.04in
    \begin{center}
    \centerline{\includegraphics[width=\columnwidth]{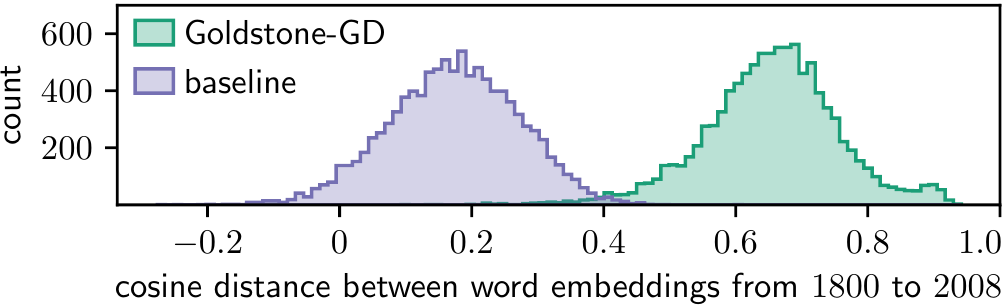}}
    \caption{Cosine distance between word embeddings from the first and last year of the training data in Dynamic Word Embeddings.}
    \label{fig:dwe-histogram}
    \end{center}
    \vskip -0.2in
\end{figure}

The focus of this paper is not the Goldstone-GD algorithm itself, but the physical intuition behind it.
We therefore limit the discussion of experiments and present only one result from~\citep{bamler2018improving}.
We applied the proposed Goldstone-GD optimization algorithm to fit Dynamic Word Embeddings~\cite{bamler_dynamic_2017}.
The model combines~$T=188$ instances of a probabilistic version of word2vec~\citep{mikolov_distributed_2013,barkan2016bayesian} with a time series prior similar to the regularizer in Eq.~\ref{eq:total-loss}.
We fitted the model to the Google Books corpus\footnote{\url{http://storage.googleapis.com/books/ngrams/books/datasetsv2.html}} \citep{michel_quantitative_2011}, following the data preparation in~\citep{bamler_dynamic_2017}.

We set ourselves a task that is very sensitive to the presence of Goldstone modes.
Given a modern query word~$w$, we want to find words~$w'$ that described the same concept in the year 1800.
To this end, we look up the embedding $z_{w,2008}$ that the fitted model assigns to~$w$ in the year 2008 (the last year in the corpus).
We then search among all embeddings in the year 1800 for the five embeddings $z_{w',1800}$ with largest cosine similarity (normalized scalar product) to $z_{w,2008}$.

Training the model without our advanced optimization algorithm lead to poor performance on this word aging task.
For example, translating the word `tuberculosis' (which was coined in 1839) from the year 2008 to the year 1800 resulted in the top five words `trained', `uniformly', `extinguished', `emerged', and `widely', which seem unrelated to the query word.
Our proposed Goldstone-GD algorithm, by contrast, provided reasonable results: `chronic', `paralysis', `irritation', `disease', and `vomiting'.
More examples of the word aging task can be found in~\citep{bamler2018improving}.

Figure~\ref{fig:dwe-histogram} provides a possible explanation for the failure of the baseline optimization method.
It shows histograms for the cosine similarity between $z_{w,2008}$ and the corresponding vectors $z_{w,1800}$ for the same word~$w$.
In the baseline (purple), no word-embeddings overlap by more than $60\%$, in strong disagreement with our prior belief that only few words change their meaning over time.
It suggests that the embedding spaces are misaligned, which is only weakly penalized if the two spaces are connected smoothly along the time axis, i.e., via a Goldstone mode.
Our proposed method does not suffer from misaligned representation spaces because it eliminates Goldstone modes.